\documentclass[runningheads]{llncs}

\usepackage{eccv}

\usepackage{eccvabbrv}

\usepackage{graphicx}
\usepackage{booktabs}

\usepackage[accsupp]{axessibility}  

\usepackage{hyperref}

\usepackage{orcidlink}

\AtBeginDocument{
\setlength{\textfloatsep}{12pt plus 6pt minus 12.0pt}
\setlength{\floatsep}{20pt plus 6pt minus 10.0pt}
\setlength{\intextsep}{12pt plus 6pt minus 6.0pt}
\setlength{\abovedisplayskip}{10.0pt plus 6pt minus 6.0pt}
\setlength{\belowdisplayskip}{10.0pt plus 6pt minus 6.0pt}
\setlength\abovedisplayshortskip{12pt plus 6pt minus 6pt}
\setlength\belowdisplayshortskip{12pt plus 6pt minus 6pt}
}

\usepackage{tabularray}
\usepackage{amssymb}
\usepackage{mathtools}
\usepackage{algpseudocode}
\usepackage{algorithm}
\usepackage{amsmath}
\usepackage{wrapfig}
\definecolor{mizuki}{HTML}{1675F2}

\def\methodName{MSSSeg}
\def\lossName{PHLoss}
\def\augName{StructAug}
\newcommand{\BC}{\text{BC}}
\newcommand{\DBC}{\text{DBC}}
\newcommand{\maxpool}{\text{MaxPool}}

\begin{document}

\title{MSSSeg: Learning Multi-Scale Structural Complexity for Self-Supervised Segmentation}

\titlerunning{MSSSeg}

\author{Haotang Li\inst{1}\orcidlink{0000-0002-1908-4801} \and
  Zhenyu Qi\inst{1}\orcidlink{0009-0003-1195-3771} \and
Hao Qin\inst{2}\orcidlink{0009-0008-2427-503X} \and Huanrui Yang\inst{1}\orcidlink{0000-0002-3384-4512} \and Kebin Peng\inst{3}\orcidlink{0000-0003-4866-786X} \and Qing Guo\inst{4}\orcidlink{0000-0003-0974-9299} \and Sen He\inst{1}\orcidlink{0000-0002-5204-8976}}

\authorrunning{Li et al.}

\institute{Department of Electrical and Computer Engineering, The University of Arizona, Tucson, AZ, USA \and
Department of Mathematics, The University of Arizona, Tucson, AZ, USA \and
Department of Computer Science, East Carolina University, Greenville, NC, USA \and VCIP, CS, Nankai University, China}

\maketitle

\begin{abstract}
Self-supervised semantic segmentation methods often suffer from structural errors, including merging distinct objects or fragmenting coherent regions, because they rely primarily on low-level appearance cues such as color and texture. These cues lack structural discriminability: they carry no information about the structural organization of a region, making it difficult to distinguish boundaries between similar-looking objects or maintain coherence within internally varying regions. Recent approaches attempt to address this by incorporating depth priors, yet remain limited by not explicitly modeling structural complexity that persists even when appearance cues are ambiguous.
To bridge this gap, we present MSSSeg, a framework that explicitly learns multi-scale structural complexity from both semantic and depth domains, via three coupled components: (1) a Differentiable Box-Counting (DBC) module that captures and aligns multi-scale structural complexity features with semantic features; (2) a Learnable Structural Augmentation (StructAug) that corrupts pixel-intensity patterns, forcing the network to rely on structural complexity features from DBC; and (3) a Persistent Homology Loss (PHLoss) that directly supervises the structural complexity of predicted segmentations. Extensive experiments demonstrate that MSSSeg achieves new state-of-the-art performance on COCO-Stuff-27, Cityscapes, and Potsdam without excessive computational overhead, validating that explicit structural complexity learning is crucial for self-supervised segmentation.
\end{abstract}
\section{Introduction}
\label{sec:intro}

Self-supervised semantic segmentation (SSS) addresses the challenge of partitioning images into semantically meaningful regions without human annotations~\cite{ijcai2022p133}, which is a capability critical for autonomous driving~\cite{vobecky2022drive}, industrial inspection~\cite{Shi_Wang_Zhang_Li_Zhu_2024}, and geographic analysis~\cite{zhang2023research}, where dense labeling is impractical.

SOTA methods~\cite{STEGO,EAGLE,ACSeg,HP} primarily build on frozen image encoders, such as DINO~\cite{caron2021emerging}, and then train segmentation heads.
These methods primarily learn from color, texture, and local patterns to group pixels into segments.
They often struggle with \emph{structural errors}, which refer to altered spatial organizations. 
\begin{figure}[ht]
    \begin{subfigure}{0.58\columnwidth}
        \centering
        \includegraphics[width=\linewidth]{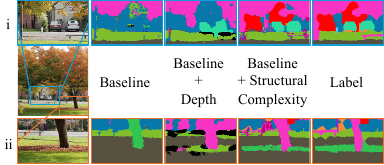}
        \caption{}
        \label{fig:structural-error}
    \end{subfigure}
    \hspace{4pt}
    \begin{subfigure}{0.40\columnwidth}
        \centering
        \includegraphics[width=\linewidth]{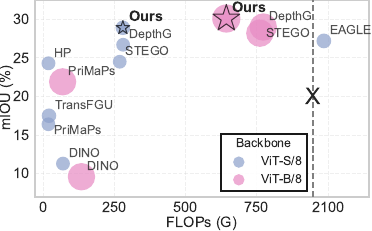} 
        \caption{}
        \label{fig:params-flops}
    \end{subfigure}
    \caption{(a) Empirical study: structural errors in SSS using COCO-Stuff-27~\cite{COCOStuff}. We use STEGO~\cite{STEGO} as Baseline. +Depth means integrating depth maps, and +Structural Complexity adds explicit supervision via our \lossName.
    (b) Performance (mIOU) vs.\ computational cost on COCO-Stuff-27. Bubble size indicates parameter count (M).
    }
    \label{fig:head}
\end{figure}
To prove this, we conduct an empirical study on the COCO-Stuff-27 dataset~\cite{COCOStuff}; we use STEGO~\cite{STEGO} as the Baseline. 
Row (i) in \Cref{fig:structural-error} shows that shadow covers the building and cars, causing Baseline to merge them into a single region. Row (ii) shows that leaves and grass have a similar texture, leading to Baseline failing to segment.
To mitigate this, some methods~\cite{hoyer2021three,DepthG} enrich feature representations by integrating depth maps, yielding partial improvements in structural coherence given that depth maps inherently encode structural complexity features that describe spatial layout rather than pixel intensity.
However, \textbf{these methods still suffer from structural errors because they do not explicitly learn structural complexity features}.
Confirmed by \Cref{fig:structural-error}, integrating depth (Baseline + Depth) partially recovers spatial layout: for row (i), cars are separated from the building, but the two cars are interconnected. In row (ii), leaves and grass are segmented, but noise from the shadow remains.
Finally, we add explicit supervision of structural complexity (Baseline + Structural Complexity) via a Persistent Homology-based loss function (Section~\ref{subsec:loss}); these errors are resolved, suggesting that explicitly learning structural complexity features is crucial.

Motivated by this, we develop \methodName, which explicitly learns structural complexity features through three coupled components without incurring prohibitive computational overhead. 
Firstly, a Differentiable Box-Counting (DBC) module extracts multi-scale structural complexity features from both pseudo-label masks and depth maps; then projects them into a shared space for cross-modal alignment. 
Secondly, a Learnable Structural Augmentation (\augName) forces the network to rely on structural complexity features from DBC. It applies learnable morphological operators to the luminance channel, corrupting local pixel intensity patterns in the input image.
Finally, we design \lossName~to ensure that structural complexity features provided by DBC are preserved. \lossName~computes persistence diagrams from the segmentation probability map, identifying connectivity defects and penalizing the critical pixels responsible.
\Cref{fig:params-flops} shows the performance (mIOU) and efficiency (FLOPs and Parameter counts) comparisons between \methodName~and SOTAs. \methodName~outperforms SOTAs on both ViT-S/8 and ViT-B/8 backbones, respectively. Notably, the performance gains are achieved without incurring excessive computational overhead.

Our main contributions are as follows:

\begin{enumerate}
    \item We present \methodName, a framework that explicitly learns structural complexity features for robust self-supervised segmentation without incurring excessive computational overhead.
    \item We introduce three coupled modules: DBC for multi-scale structural complexity feature extraction and cross-modal alignment, a learnable \augName~for adversarial structural corruption, and a \lossName~for sparse structure-level supervision at critical pixels.
    \item We achieve new state-of-the-art performance on three benchmarks (COCO-Stuff-27, Cityscapes, and Potsdam), and provide comprehensive ablation studies that validate each component's contribution.
\end{enumerate}

\section{Related Work}
\label{sec:related_work}

\textbf{Semantic Segmentation: }
Following the success of self-supervised vision transformers like DINO~\cite{DINOv2}, a dominant paradigm has emerged: freezing the pre-trained backbone and training lightweight, task-specific segmentation heads.
For example, Hamilton \etal \cite{STEGO} distilling feature correspondences from pre-trained self-supervised vision transformers to form consistent, discrete semantic labels.
Yin \etal \cite{TransFGU} propose a top-down framework by extracting high-level semantic priors from self-supervised models and mapping them to low-level pixel features via class activation maps.
Seong \etal \cite{HP} conduct contrastive learning by mining global and local hidden positives to capture rich semantic relationships and ensure spatial consistency.
Hahn \etal \cite{PriMaPs} decompose images into semantically meaningful masks based on their self-supervised feature representations, fitting class prototypes via a stochastic expectation-maximization algorithm.
Li \etal \cite{ACSeg} encode concepts into learnable prototypes for pixel-level semantic aggregation in self-supervised vision transformer pre-trained models for segmentation.
EAGLE \cite{EAGLE} introduces EiCue to provide semantic and structural cues through eigen vectors derived from the semantic similarity matrix of image features and color affinity from images.

\noindent\textbf{Auxiliary Depth Information in Semantic Segmentation: }
Most self-supervised segmentation methods operate solely on features derived from RGB images, relying on pixel-level similarity for grouping. To move beyond pure appearance, recent works have explored auxiliary cues that encode additional scene information. 
Sick \etal \cite{DepthG} integrate 3D scene structure through depth-guided feature correlation and utilize farthest-point sampling to effectively select spatially relevant features.
Hoyer \etal \cite{hoyer2021three} leverage depth estimation to transfer learned depth features, perform geometry-aware data augmentation via DepthMix, and automatically select the most informative samples for annotation.
Wang \etal \cite{wang2021domain} bridge the domain gap by leveraging depth estimation to explicitly learn cross-modal feature correlations and refine target pseudo-labels based on depth prediction discrepancies.
\section{\methodName}
\label{sec:approach}

\begin{figure*}[ht]
    \centering
    \includegraphics[width=\linewidth]{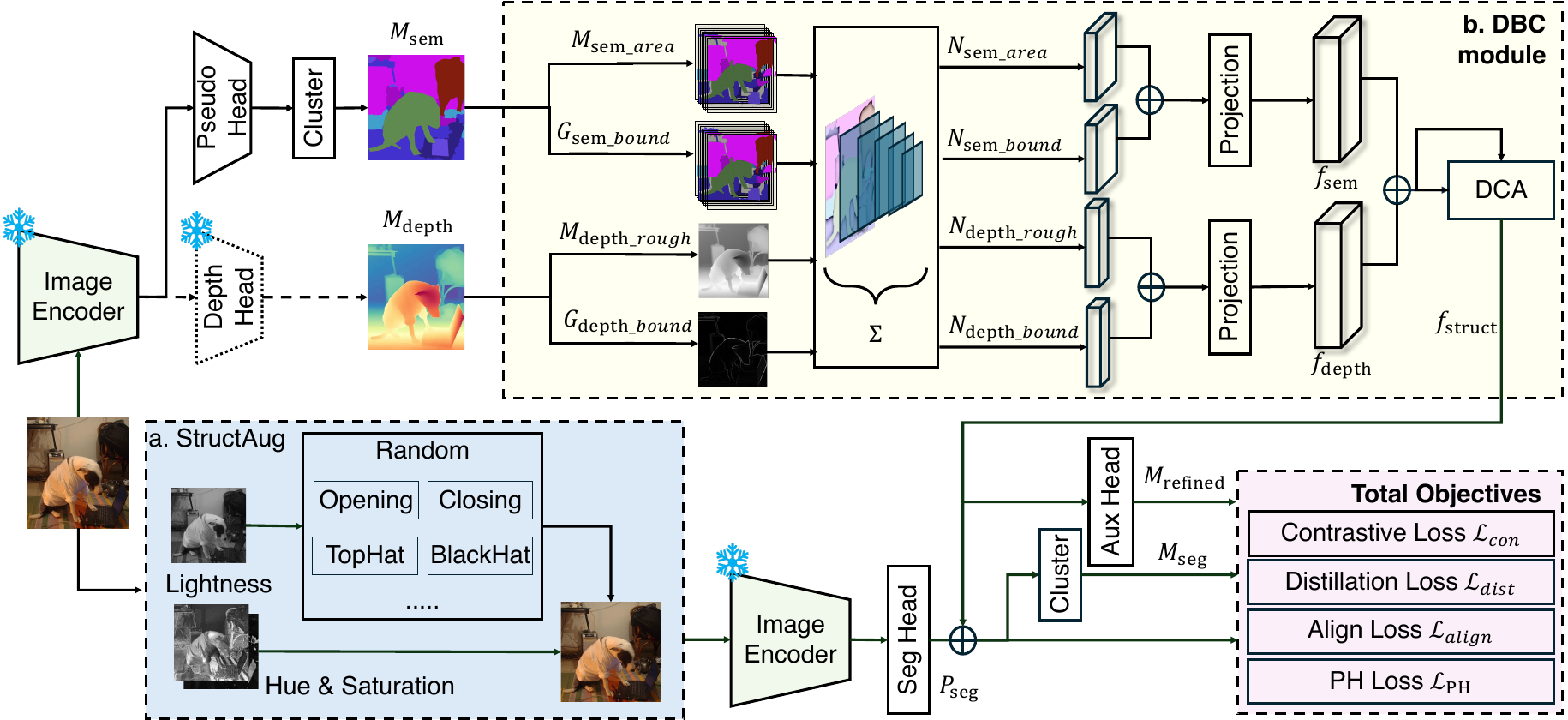} 
    \caption{Overview of \methodName. The framework introduces three components into the standard SSS pipeline: (a) a DBC module that explicitly extracts multi-scale structural complexity from both RGB images and depth maps, (b) StructAug that forces the network to rely on DBC features, and (c) a multi-objective \lossName~that preserves structural complexity in the output. Both training and inference pipelines are shown.}
    \label{fig:overview}
\end{figure*}

\subsection{Overall Framework}
Most existing self-supervised segmentation methods \cite{STEGO,EAGLE,ACSeg,HP} share a common pipeline.
Given an input image $I$, a frozen encoder $\mathcal{E}$ extracts pixel-level features, a clustering step produces pseudo-labels $M_\text{sem} = \text{Cluster}(\mathcal{E}(I))$, and a segmentation head is trained on an augmented view under pixel-wise supervision (Equation~\ref{eq:baseline}):
\begin{equation}
    M_\text{seg} = \text{SegHead}\big(\mathcal{E}(\text{Aug}(I)),\; M_\text{sem},\; \mathcal{A}(I)\big), \quad \mathcal{L} = \mathcal{L}_\text{pixel}
    \label{eq:baseline}
\end{equation}
where $\text{Aug}$ denotes standard photometric augmentation, $\mathcal{A}(I)$ represents optional auxiliary information, and $\mathcal{L}_\text{pixel}$ denotes pixel-wise losses.
Recent depth-fusion methods~\cite{DepthG} partially alleviate structural errors by supplying depth maps through $ \mathcal{A}(I) $, yet they integrate depth at the raw feature level \textit{without explicitly extracting its structural complexity features.} As shown in \Cref{fig:structural-error}, fusion depth maps are insufficient to resolve structural errors such as fragmentation or merging caused by shadows and textures.

To explicitly extract and align multi-scale structural complexity features from both pseudo-label masks and depth maps, \methodName~introduces three components into this pipeline, illustrated in Equation~\ref{eq:ours} and shown in \Cref{fig:overview}:
\begin{align}
    \label{eq:ours}
    M_\text{seg} = \text{SegHead}\big(\mathcal{E}(\text{StructAug}_\theta(I)),\; [f_\text{sem};\, f_\text{depth}] \big),& \\
    \mathcal{L} = \mathcal{L}_\text{pixel} + \mathcal{L}_\text{align} + \mathcal{L}_\text{PH}& \nonumber
\end{align}
The \DBC~module (\Cref{subsec:dbc}) extracts multi-scale structural complexity features from both the pseudo-label mask $M_\text{sem}$ and a depth map $\mathcal{D}(I)$, projecting them into aligned vectors $[f_\text{sem};\, f_\text{depth}] = \text{DBC}(M_\text{sem},\, \mathcal{D}(I))$.
\augName~(\Cref{subsec:structaug}) corrupts the luminance channel of $I$ so that the Seg Head, observing only the corrupted view, cannot rely on low-level pixel-intensity cues and must depend on the structural complexity features from DBC.
A Deformable Cross Attention (DCA)~\cite{DCA} decoder fuses $f_\text{struct} = [f_\text{sem};\, f_\text{depth}]$ with the encoder features $\mathcal{E}(I)$, producing a refined prediction $M_\text{refined}$.
Unlike $P_\text{seg}$, which is produced from corrupted features alone $ P_\text{seg} = \text{SegHead}\big(\mathcal{E}(\text{StructAug}_\theta(I))) $, $M_\text{refined}$ integrates explicit structural complexity knowledge from both semantic and depth domains, yielding a prediction that is structurally coherent even where pixel-intensity cues are ambiguous.
$M_\text{refined}$ is directly supervised by $\mathcal{L}_\text{pixel}$, an alignment loss $\mathcal{L}_\text{align}$ forces $f_\text{sem}$ and $f_\text{depth}$ to encode consistent structural descriptions, and \lossName~($\mathcal{L}_\text{PH}$, \Cref{subsec:loss}) penalizes connectivity defects in the Seg Head output at their critical pixels, ensuring structural complexity is preserved, total objectives are detailed in \Cref{subsec:total-loss}.

At inference, StructAug is repurposed as structure-aware test-time augmentation by generating $N$ perturbed views and averaging their softmax maps before this fusion step. Seg Head produces $M_\text{seg}$ from the input image. $M_\text{seg}$ and $\mathcal{D}(I)$ are then routed through the DBC module, and the DCA decoder produces $M_\text{refined}$. The two predictions are fused before clustering and DenseCRF post-processing. 

\subsection{Differentiable Box Counting Module}
\label{subsec:dbc}

The structural errors occur because pixel-level features cannot capture how regions connect, fragment, and vary in structural complexity across scales. To provide the network with this missing information, we need a descriptor that explicitly measures structural complexity at multiple resolutions. The box counting dimension from fractal geometry~\cite{mandelbrot1982fractal} formalizes exactly this: it quantifies how the spatial occupancy of a set changes with scale, producing a single value sensitive to connectivity, boundary complexity, and fragmentation.

Given a feature map $M \in \mathbb{R}^{B \times C \times H \times W}$ and a box size $s$, standard box counting partitions the spatial domain into $s \times s$ boxes, $B_s(i,j)$, and counts how many contain nonzero contents (\Cref{eq:bc}):
\begin{align}
    \BC(M, s) \coloneqq \textstyle \sum_{i, j} \mathbb{I} \bigl( \sup_{x \in B_{s}(i,j)} M(x) > 0 \bigr)
    \label{eq:bc}
\end{align}
The indicator function $\mathbb{I}(\cdot)$ blocks gradient flow, so we replace it with the continuous occupancy signal, implemented as a sum of max pool outputs (\Cref{eq:dbc}):
\begin{align}
    \DBC(M, s) \coloneqq \textstyle \sum_{i, j} \sup_{x \in B_{s}(i,j)} M(x)
    \label{eq:dbc}
\end{align}
Evaluating $\DBC$ at geometrically spaced box sizes $\mathbf{S} = \{s_1, \ldots, s_k\}$ produces a $k$ dimensional structural complexity signature (\Cref{eq:ddbc}):
\begin{align}
    \mathbf{DBC}(M, \mathbf{S}) \coloneq \textstyle \bigl( \DBC(M, s) \bigr)_{s \in \mathbf{S}}
    \label{eq:ddbc}
\end{align}
Coarse scales (large-sized boxes) in this signature reflect the number and spread of connected components, while fine scales (small-sized boxes) capture boundary irregularity and local fragmentation. Together they form a complete complexity profile of $M$ across resolutions.

Depth maps inherently encode structural complexity rather than pixel intensity: depth discontinuities reveal where objects physically separate, and surface variations capture how regions are spatially arranged.
Yet SOTA methods~\cite{DepthG,hoyer2021three} fuse depth maps at the feature level without explicitly extracting this information. We first apply $\mathbf{DBC}$ to depth maps to obtain explicit structural complexity features that these methods miss, and then apply $\mathbf{DBC}$ to the pseudo-label segmentation to align the two. Each domain is probed through an area-view and a boundary-view, yielding four descriptors whose combination is validated via ablation in \Cref{subsec:ablation}. The structure shows in \Cref{fig:overview}-b.

From a normalized depth map $M_\text{depth} \in \mathbb{R}^{B \times H \times W}$, we extract two descriptors that capture the scene's physical structure at multiple scales.
\emph{Surface roughness} measures how much the depth surface varies locally (e.g., distinguishing a cobblestone road from a smooth wall).
We compute a local variance map $M_\text{depth\_var} = \mathbb{E}_{3 \times 3}[M_\text{depth}^2] - \mathbb{E}_{3 \times 3}[M_\text{depth}]^2$, where $\mathbb{E}_{3 \times 3}[\cdot]$ is a $3 \times 3$ sliding window mean, and apply
$N_\text{depth\_rough} = \mathbf{DBC}(M_\text{depth\_var}, \mathbf{S})$.
\emph{Depth edges} capture depth discontinuities that typically coincide with object boundaries.
We compute the gradient magnitude $G_\text{depth}$ with a Sobel filter and apply
$N_\text{depth\_bound} = \mathbf{DBC}(G_\text{depth}, \mathbf{S})$.
Together, these two descriptors provide an explicit, multi-scale characterization of the scene's physical layout that is independent of pixel intensity.

To align the segmentation with the depth structure above, we also need to measure the structural complexity of what the network currently believes. We start from the pseudo label map $M_\text{sem} \in \mathbb{R}^{B \times H \times W}$ and convert it into per class binary masks $M_\text{sem} \in \{0,1\}^{B \times C \times H \times W}$ via one hot encoding.
\emph{Area counting} applies $\mathbf{DBC}$ directly to $M_\text{sem}$, measuring how each class region fragments across scales:
$N_\text{sem\_area} = \mathbf{DBC}(M_\text{sem}, \mathbf{S})$.
\emph{Boundary counting} first isolates class boundaries via the morphological gradient $G_\text{seg} = \text{Dilation}(M_\text{sem}) - \text{Erosion}(M_\text{sem})$, then measures their complexity:
$N_\text{sem\_bound} = \mathbf{DBC}(G_\text{seg}, \mathbf{S})$.
$N_\text{sem\_area}$ captures \emph{what} is segmented (region connectivity and dispersion) while $N_\text{sem\_bound}$ captures \emph{how} it is segmented (boundary regularity).

The four counting tensors live in different statistical regimes: semantic counts are discrete-valued over class masks, whereas depth counts are continuous-valued over depth derivatives. Concatenating them directly would let one domain dominate gradient magnitudes.
We therefore project each domain through its own lightweight Multi-layer Perceptron (MLP): depth tensors $[N_\text{depth\_rough};\, N_\text{depth\_bound}]$ are mapped to $f_\text{depth} \in \mathbb{R}^{B \times D}$, and semantic tensors $[N_\text{sem\_area};\, N_\text{sem\_bound}]$ to $f_\text{sem} \in \mathbb{R}^{B \times D}$.
Because both vectors now live in the same $D$ dimensional space, we can directly measure their agreement with a cosine similarity loss~\cite{CELoss} $\mathcal{L}_\text{align} = 1 - \cos(f_\text{sem},\, f_\text{depth})$, which forces the segmentation's structural complexity to be consistent with the scene's physical layout. The resulting vectors are concatenated into $f_\text{struct} = [f_\text{sem};\, f_\text{depth}] \in \mathbb{R}^{B \times 2D}$ and injected into the DCA decoder to produce $M_\text{refined}$.

\subsection{Learnable Structural Augmentation}
\label{subsec:structaug}

Standard photometric augmentations~\cite{photometric} (ColorJitter, GaussianBlur) perturb global statistics; however, they cannot synthesize \emph{local structural deformation characteristics} of real illumination artifacts. For example, shadows and textures can incorrectly split or merge regions, which are fundamentally structural errors.
To address this, we introduce Learnable Structural Augmentation (StructAug), a differentiable module whose corruption behavior is \emph{jointly optimized} with the segmentation network, parameterizing operator selection, boundary sensitivity, and corruption intensity with learnable parameters updated end-to-end.

To avoid non-physical color artifacts from operating on RGB directly~\cite{sun2023order}, we convert the input to HSL and restrict all operations to the Lightness channel $L$~\cite{zhang2024color}, preserving semantic-bearing Hue and Saturation.

We replace the fixed hyperparameters in classical erosion--dilation~\cite{morph} with learnable counterparts.
Given $L$ and a local neighborhood $\mathcal{N}_k(x)$, the \emph{Learnable Geodesic Erosion} and \emph{Dilation} are \Cref{eq:learn_erosion} and \Cref{eq:learn_dilation}, respectively:
\begin{align}
    \varepsilon_\theta(L)(x) &= \min_{y \in \mathcal{N}_k(x)} \Big[ L(y) + \bigl(1 - w_\theta(x, y)\bigr) \cdot \alpha_\varepsilon \Big] \label{eq:learn_erosion} \\
    \delta_\theta(L)(x) &= \max_{y \in \mathcal{N}_k(x)} \Big[ L(y) - \bigl(1 - w_\theta(x, y)\bigr) \cdot \alpha_\delta \Big] \label{eq:learn_dilation}
\end{align}
where the learnable affinity kernel $w_\theta(x, y) = \exp\!\bigl( -{|L(x) - L(y)|^2}/{2\sigma^2_\theta} \bigr)$ governs boundary sensitivity.
The per-operator parameter $\sigma_\theta \in \mathbb{R}^+$ controls how aggressively the operator crosses intensity edges, while the learnable penalty magnitudes $\alpha_\varepsilon, \alpha_\delta \in \mathbb{R}^+$ independently control the corruption strength of erosion and dilation. All are parameterized via softplus~\cite{softplus}.
This construction continuously interpolates between isotropic flat morphology ($\sigma_\theta \to \infty$) and boundary-preserving geodesic morphology ($\sigma_\theta \to 0$), learning the optimal regime from data. 

Composing these primitives yields a toolbox $\mathcal{T}$: Geodesic Opening $\mathcal{O}_\theta = \delta_\theta \circ \varepsilon_\theta$ (breaks bright bridges, simulating specular suppression), Geodesic Closing $\mathcal{C}_\theta = \varepsilon_\theta \circ \delta_\theta$ (fills dark gaps, simulating cast shadows), Black Top-Hat $\mathcal{B}_\theta = \mathcal{C}_\theta - \text{id}$ (isolates shadow residuals), and White Top-Hat $\mathcal{W}_\theta = \text{id} - \mathcal{O}_\theta$ (isolates glare residuals).
Each operator $T_i \in \mathcal{T}$ is associated with a learnable logit $\ell_i$; selection uses Gumbel-Softmax~\cite{jang2017categorical,maddison2017concrete} sampling $\hat{T} = \sum_i \tilde{p}_i \cdot T_i(L)$, enabling discrete selection with differentiable gradient updates.
A learnable blending coefficient $\beta = \text{sigmoid}(\ell_\beta)$ controls the final mixing in~\Cref{eq:blend}:
\begin{equation}
    L_{\text{aug}} = (1 - \beta)\, L + \beta\, \hat{T}
    \label{eq:blend}
\end{equation}
The augmented $L_{\text{aug}}$ is recombined with the original H and S channels and converted back to RGB.
The full parameter set $\theta = \{\sigma_\varepsilon, \sigma_\delta, \alpha_\varepsilon, \alpha_\delta, \{\ell_i\}, \ell_\beta\}$ is compact (fewer than 10 scalars) yet controls what corruption to apply, how severely it deforms structure, and how strongly it is blended.
All parameters receive gradients through the segmentation loss, so the augmentation co-evolves with the model: it naturally intensifies as the segmentation head strengthens, producing an adversarial curriculum without manual annealing. We perform an ablation study on \augName~on \Cref{subsec:ablation}.

\subsection{Persistent Homology Loss}
\label{subsec:loss}
Pixel-wise objectives such as cross-entropy~\cite{CELoss} and Dice loss\cite{Dice} evaluate each spatial location independently, providing no mechanism to penalize \emph{structural} errors.
For example, a predicted mask may score high per pixel accuracy yet contain broken connections or spurious holes that alter the spatial organization of the segmentation.
These errors are especially damaging in our setting: because the DBC module explicitly extracts structural complexity features (connected-component counts, boundary complexity) to guide learning, the segmentation head must produce masks whose global connectivity faithfully reflects the scene structure.
To close this gap, we introduce a \emph{Persistent Homology} (PH) loss, $\mathcal{L}_\text{PH}$, directly supervises the structure complexity of the predicted probability map.

Rather than thresholding the Seg Head's continuous output $P_\text{seg} \in [0,1]^{H \times W}$ into a hard mask, we treat it as a structural landscape and analyze it via \emph{superlevel-set filtration}~\cite{clough2020topological}.
Conceptually, a threshold $v$ sweeps from $1$ down to $0$; at each level, the superlevel set $\{x \mid P_\text{seg}(x) \geq v\}$ defines a binary mask whose structural complexity evolves continuously.
During this sweep, structural complexity features undergo two types of critical events.
A \emph{birth} occurs at threshold $v = b$ when a new connected component ($\beta_0$) emerges at a local maximum, or a new hole ($\beta_1$) forms as a loop closes.
A \emph{death} occurs at $v = d$ when two components merge at a saddle point (the younger one dies) or a hole is filled.
Each feature is thus recorded as a birth--death pair $(b_i, d_i)$, where $b_i = P_\text{seg}(c_{\text{birth}}^{(i)})$ and $d_i = P_\text{seg}(c_{\text{death}}^{(i)})$ are the network's predicted probabilities at the specific critical pixels, $c_{\text{birth}}$ and $c_{\text{death}}$, that create and destroy the feature.
The collection of all such pairs forms the \emph{Persistence Diagram} $\text{PD}(P_\text{seg})$, which is a complete summary of the map's structure complexity: features far from the diagonal $b = d$ represent prominent, long-lived structures, while features near the diagonal correspond to structural noise.

To define a supervisory signal, we compute a reference diagram $\text{PD}(M_\text{sem})$ from the pseudo-label mask and establish a one-to-one correspondence between the two diagrams via optimal bipartite matching (Wasserstein distance~\cite{Wasserstein}).
This matching partitions the predicted features into two disjoint sets: \emph{true signals} $\mathcal{S}$ (features matched to a real structure in the reference) and \emph{structural noise} $\mathcal{N}$ (spurious features with no counterpart).
The PH loss applies a distinct penalty to each set:
\begin{align}
\mathcal{L}_\text{true} &= \textstyle \sum_{i \in \mathcal{S}} \bigl[(1 - b_i)^2 + d_i^{\,2}\bigr] \label{eq:ph_true} \\
\mathcal{L}_\text{noise} &= \textstyle \sum_{j \in \mathcal{N}} (b_j - d_j)^2 \label{eq:ph_noise}
\end{align}
For true signals, \Cref{eq:ph_true} drives each matched structure toward maximum prominence by pushing its birth probability to $1$ (fully confident foreground) and its death probability to $0$ (persisting to the background level).
For structural noise, \Cref{eq:ph_noise} collapses each spurious feature onto the diagonal $b = d$, annihilating it by forcing its birth and death to coincide so the structure never exists.
The total PH loss is $\mathcal{L}_\text{PH} = \mathcal{L}_\text{true} + \mathcal{L}_\text{noise}$.

A distinctive property of $\mathcal{L}_\text{PH}$ is its \emph{gradient sparsity}.
Because $b_i$ and $d_i$ are the network's output at specific critical pixels, the gradients of the loss are exactly zero at all non-critical locations.
Even for a dense $H \times W$ probability map, backpropagation through $\mathcal{L}_\text{PH}$ only updates the handful of pixels (local extrema and saddle points) that are directly responsible for creating or destroying structural complexity features.
This acts as a highly targeted surgical correction: it fixes the exact pixels causing a structural break or a spurious hole without disturbing the rest of the prediction, complementing the dense, pixel-level gradients from auxiliary objectives (detailed in \Cref{subsec:total-loss}) that handle semantic accuracy.

\subsection{Total Objective}
\label{subsec:total-loss}
We jointly optimize \methodName~with the multi-objective loss $\mathcal{L}$, which pairs the Persistent Homology loss $\mathcal{L}_\text{PH}$ introduced in \Cref{subsec:loss} with four auxiliary objectives.
A pixel-prototype contrastive loss $ \mathcal{L}_\text{con} $, defined by an InfoNCE~\cite{InfoNCE} objective, pulls each pixel feature $ \mathbf{e}_i $ from the Seg Head embedding $ \mathbf{E} $ toward its corresponding class prototype $ \hat{\mathbf{e}}_i $ from the Pseudo Head while pushing it away from all other prototypes, ensuring the feature space is compact and separable.
A distillation loss~\cite{CELoss} $ \mathcal{L}_\text{dist} $ applies cross-entropy between the Seg Head's softmax probability map $ P_\text{seg} = \sigma(\phi(\mathbf{E})) $ and the stop-gradiented pseudo-label $ \text{sg}(M_\text{sem}) $, mapping the learned features to correct semantic classes.
We group $ \mathcal{L}_\text{con} $ and $ \mathcal{L}_\text{dist} $ as the pixel level loss $ \mathcal{L}_\text{pixel} = \mathcal{L}_\text{con} + \mathcal{L}_\text{dist} $, as both supervise semantic accuracy rather than structural organization,
 combine as pixel loss $ \mathcal{L}_\text{pixel} $.
A cross-modal structure alignment loss~\cite{CosineSim} $ \mathcal{L}_\text{align} = 1 - \text{cos}(f_\text{sem}, f_\text{depth}) $ forces the semantic and depth structural complexity vectors from the DBC module to align, ensuring cross-modal consistency.
The sparse, structure-level gradients from $\mathcal{L}_\text{PH}$ correct structural complexity defects at critical pixels, while the dense, pixel-level supervision from the auxiliary terms ensures semantic accuracy, together producing segmentation that is structurally consistent. 
The total objective is shown in \Cref{eq:loss}, where $\lambda$s are weighting coefficients.
\begin{equation}
    \label{eq:loss}
\mathcal{L} = \lambda_\text{con} \mathcal{L}_\text{con} + \lambda_\text{dist} \mathcal{L}_\text{dist} + \lambda_\text{align} \mathcal{L}_\text{align} + \lambda_\text{PH} \mathcal{L}_\text{PH}
\end{equation}
\section{Evaluation}
\label{sec:evaluation}

\subsection{Experiment Setup}
\label{subsec:experiment-setup}

\textbf{Implementation Details}
We use DINO \cite{DINOv2} pretrained vision transformers (ViT-S/8 and ViT-B/8, respectively) as our baseline Image Encoders, which are kept frozen during the training process. Following prior work \cite{EAGLE,STEGO}, a frozen depth estimator pretrained on DINO \cite{DepthAnythingV2} is used to provide depth priors.
The training images are resized and randomly cropped to $ 224 \times 224 $.

For the evaluation, we also follow the protocols used in prior work \cite{STEGO,DepthG}. While it generates class-agnostic clusters, we use the Hungarian matching algorithm~\cite{HunMat} to find the optimal one-to-one mapping between our predictions and the ground-truth classes for a fair evaluation. 

\textbf{Datasets and Metrics}
We evaluate our framework on three diverse and challenging benchmarks: (1) COCO-Stuff \cite{COCOStuff} which has detailed pixel-level annotations, facilitating comprehensive various object understanding, we evaluate on the 27 mid-level categories; (2) Cityscapes \cite{Cityscapes}, we evaluate on the 27 foreground classes, which is a common setup used in recent unsupervised/self-supervised segmentation literature for comprehensive scene parsing; (3) Potsdam \cite{Potsdam} originally consisting of 6 classes, is merged into 3 distinct super-categories for this task: "Buildings", "Vegetation" (merging "Tree" and "Low veg."), and "Ground" (merging "Impervious surfaces", "Car", and "Clutter").
For qualitative evaluation, we adopt mean intersection over union (mIoU) and pixel accuracy (Acc) as metrics, following SOTA research on semantic segmentation. We show additional results on the Pascal VOC 2012 dataset \cite{PASCALVOC2012} in Supplemental Materials.

\begin{table*}[ht]
\footnotesize
\centering
\caption{Quantitative comparison with SOTA SSS methods across three standard benchmarks: COCO-Stuff, Cityscapes, and Potsdam. Results are reported for both ViT-S/8 and ViT-B/8 backbones using Acc and mIoU. "-" represents no result reported by the related paper.}
\label{tab:quantitative}
\begin{tblr}
    {rowsep=0.5pt, 
    width=\linewidth, 
    colspec={X[2.5,l,m]X[1.8,c,m]X[1,c,m]X[1,c,m]X[1,c,m]X[1,c,m]X[1,c,m]X[1,c,m]},
    hline{2} = {3-4}{leftpos = -0.5, rightpos = -0.5, endpos},
    hline{2} = {5-6}{leftpos = -0.5, rightpos = -0.5, endpos},
    hline{2} = {7-8}{leftpos = -0.5, rightpos = -0.5, endpos},
    hline{2} = {9}{leftpos = -0.5, rightpos = -0.5, endpos},
    row{10,17} = {gray9},
    }
\hline
\SetCell[r=2]{} Method & \SetCell[r=2]{} Backbone & \SetCell[c=2]{} COCO-Stuff &  & \SetCell[c=2]{} Cityscapes &  & \SetCell[c=2]{} Potsdam &  \\
 &  & Acc & mIoU & Acc & mIoU & Acc & mIoU \\
\hline
DINO \cite{DINOv2} & ViT-S/8 & 28.7 & 11.3 & 34.5 & 10.9 & 56.6 & 33.6 \\
STEGO \cite{STEGO} & ViT-S/8 & 48.3 & 24.5 & - & - & - & - \\
TransFGU \cite{TransFGU} & ViT-S/8 & 52.7 & 17.5 & 77.9 & 16.8 & - & - \\
HP \cite{HP} & ViT-S/8 & 54.5 & 24.3 & 80.1 & 18.4 & - & - \\
PriMaPs \cite{PriMaPs} & ViT-S/8 & 46.5 & 16.4 & 81.2 & 19.4 & 62.5 & 38.9 \\
DepthG \cite{DepthG} & ViT-S/8 & 55.1 & 26.7 & - & - & 80.4 & - \\
EAGLE \cite{EAGLE} & ViT-S/8 & 64.2 & 27.2 & 81.8 & 19.7 & - & - \\
\textbf{Ours} & ViT-S/8 & \textbf{71.1} & \textbf{30.2} & \textbf{85.3} & \textbf{22.1} & \textbf{86.4} & \textbf{70.6} \\
\hline
DINO \cite{DINOv2} & ViT-B/8 & 30.5 & 9.6 & 43.6 & 11.8 & 66.1 & 49.4 \\
STEGO \cite{STEGO} & ViT-B/8 & 56.9 & 28.2 & 73.2 & 21.0 & 77.0 & 62.6 \\
HP \cite{HP} & ViT-B/8 & - & - & 79.5 & 18.4 & 82.4 & 68.6 \\
PriMaPs \cite{PriMaPs} & ViT-B/8 & 48.5 & 21.9 & 59.6 & 17.6 & 80.5 & 67.0 \\
EAGLE \cite{EAGLE} & ViT-B/8 & - & - & 79.4 & 22.1 & 83.3 & 71.1 \\
DepthG \cite{DepthG} & ViT-B/8 & 58.6 & 29.0 & 81.6 & 23.1 & - & - \\
\textbf{Ours} & ViT-B/8 & \textbf{73.5} & \textbf{34.9} & \textbf{87.4} & \textbf{26.8} & \textbf{89.2} & \textbf{75.7} \\

\hline
\end{tblr}
\end{table*}

\subsection{Quantitative Analysis}
\label{subsec:quantitative}
We present a comprehensive comparison of our \methodName~framework against the leading SOTA methods in self-supervised semantic segmentation.
The quantitative results, presented in \Cref{tab:quantitative}, demonstrate that our method achieves a new SOTA performance across all three challenging benchmarks on both metrics.

Analyzing the ViT-S/8 results, on the COCO-Stuff-27, our 71.1 Acc and 30.2 mIoU surpass the previous best method, EAGLE (64.2 Acc and 27.2 mIoU). 
This lead continues on Cityscapes, where our 85.3 Acc and 22.1 mIoU represent a +3.5 Acc and a +2.4 mIoU leap over the next-best competitor, EAGLE. This demonstrates the crucialness of learning structural complexity features in SSS. 
Furthermore, the performance on the Potsdam dataset achieves 86.4 Acc and 70.6 mIoU, outperforming the next-best reported competitors, DepthG (80.4 Acc) and PriMaPs (38.9 mIoU). Given that Potsdam is dominated by structures such as buildings and roads, where semantic color cues are often unreliable, this significant margin strongly supports that our model's success is directly attributable to the explicit modeling of structural complexity features.

We also evaluate performance on a larger ViT-B/8 backbone. \methodName~achieves the best Acc and mIoU results across all three datasets, including 73.5 Acc (+14.9) and 34.9 mIoU (+5.9) on COCO-Stuff; 87.4 Acc (+5.8) and 26.8 mIoU (+3.7) on Cityscapes, when comparing to the second best reported SOTA, depth-fusion method: DepthG. The performance gains support the effectiveness of explicitly learning structural complexity features in SSS.
Finally, in Potsdam \methodName~shows 89.2 Acc (+5.9) and 75.7 mIoU (+4.6), compared with EAGLE. 

Across two backbones on three diverse datasets, our framework, by leveraging multi-scale structural complexity, consistently raises the bar for SSS.

\subsection{Qualitative Analysis}
\label{subsec:qualitative}

\begin{figure*}[ht]
    \centering
    \includegraphics[width=\linewidth]{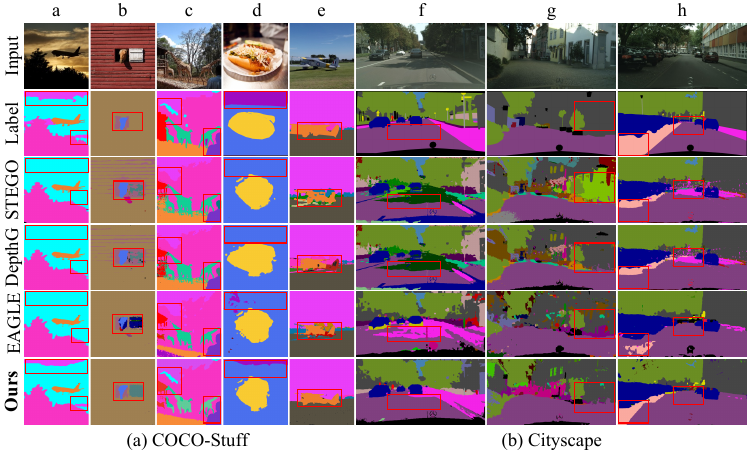}
    \caption{Qualitative semantic segmentation comparisons on the (a) COCO-Stuff-27 \cite{COCOStuff} (a - e) and (b) Cityscapes \cite{Cityscapes} (f - h) datasets. \methodName~(ViT/S-8 version) consistently produces more spatially coherent and accurate masks compared to SOTAs (STEGO, DepthG, and EAGLE).}
    \label{fig:qualitative}
\end{figure*}

Beyond the quantitative gains demonstrated in \Cref{subsec:quantitative}, we provide qualitative comparisons in \Cref{fig:qualitative} to visually demonstrate our framework's (ViT/S-8) superiority over the leading baselines, STEGO~\cite{STEGO}, DepthG~\cite{DepthG} and EAGLE~\cite{EAGLE}.

Across a variety of challenging scenes from COCO-Stuff-27, our method (\textbf{Ours}) produces segmentation masks that are more coherent and spatially precise to the groundtruth labels (Label). 
This is particularly evident in \Cref{fig:qualitative}-a and \Cref{fig:qualitative}-c, where all three compared SOTA methods fail to segment the cloud in the sky.
In contrast, our method, guided by explicitly learned structural complexity features from depth maps, successfully distinguishes cloud areas.
Also, the baselines fail to separate distinct adjacent objects, such as merging the background and table (\Cref{fig:qualitative}-d) into single, incorrect semantic blobs.
Furthermore, without explicitly structural complexity features learned from depth maps, results from all other SOTAs are degraded by shadows (\Cref{fig:qualitative}-e).

This robust performance extends to complex street scenes, as seen in the Cityscapes examples, where our model demonstrates a clear superiority in parsing complex structures in \Cref{fig:qualitative}-f and \Cref{fig:qualitative}-h, where all three compared methods fail to capture the immediate road surface, shattering it into multiple spurious regions.
Our model, in contrast, clearly identifies the road as a continuous object.
Furthermore, all baselines fail to capture the building's facade in \Cref{fig:qualitative}-g, shattering it into multiple spurious regions and hallucinating a "sky" class, our \methodName~correctly identifies the entire "building" as a single, unified object.

This strong visual evidence confirms that our quantitative Acc and mIoU lead is the direct result of a fundamentally more robust approach that benefits from explicitly learning structural complexity features.

\subsection{Ablation Study}
\label{subsec:ablation}
To validate our framework and isolate the effectiveness of each component, we conduct a series of comprehensive ablation studies.
All ablation experiments are conducted on the challenging COCO-Stuff-27 benchmark, using the DINO-pretrained ViT-S/8 backbone. We report the standard Acc and mIoU metrics.

\begin{table}[ht]
\centering
\caption{Ablation study on the core components, evaluated on COCO-Stuff. Exp. i and vi validate the necessity of \augName. Exp. ii - vi demonstrates that all four structural complexity descriptors in the DBC module, Semantic Area (SA), Semantic Boundary (SB), Depth Area (DA), and Depth Boundary (DB).}
\label{tab:ablation-framework}
\begin{tblr}{rowsep=0.5pt, width=\columnwidth, colspec={X[1,c,m]X[1,c,m]X[1,c,m]X[1,c,m]X[1,c,m]X[1,c,m]|X[1,c,m]X[1,c,m]}}
\hline
\SetCell[r=2]{} Exp. & \SetCell[r=2]{} \augName & \SetCell[c=4]{} Box-Counting &  &  &  & \SetCell[r=2]{} Acc & \SetCell[r=2]{} mIoU \\
 &  & SA & SB & DA & DB &  &  \\
\hline
i &  & \checkmark & \checkmark & \checkmark & \checkmark & 67.9 & 28.7 \\
ii & \checkmark &  & \checkmark & \checkmark & \checkmark & 69.9 & 29.5 \\
iii & \checkmark & \checkmark &  & \checkmark & \checkmark & 70.8 & 30.0 \\
iv & \checkmark & \checkmark & \checkmark &  & \checkmark & 69.4 & 29.2 \\
v & \checkmark & \checkmark & \checkmark & \checkmark &  & 68.8 & 29.1 \\
vi & \checkmark & \checkmark & \checkmark & \checkmark & \checkmark & 71.1 & 30.2 \\
\hline
\end{tblr}
\end{table}

We first validate the necessity of \augName~and design of the DBC module, with results presented in \Cref{tab:ablation-framework}.
The \checkmark denotes applying \augName~to the segmentation branch.
For the \augName, row i and row vi confirm that \augName~act as an adversarial augmentation, creating a crucial information gap that forces it to learn robust structural complexity features. Without \augName, the model has -3.2 Acc and -1.5 mIOU drop.
Next, for the internal design of the DBC Module, which is built on four structural complexity descriptors: Semantic Area (SA), Semantic Boundary (SB), Depth Area (DA), and Depth Boundary (DB).
Starting from our full-version Model (30.2 mIoU), we ablate each component, as shown in rows ii-v of \Cref{tab:ablation-framework}. Removing any single component results in a performance drop, confirming that they are all necessary.
Specifically, removing SA reduces the mIoU to 29.5, and removing SB reduces it to 30.0.
The most significant degradation occurs when removing DA and DB, which reduces performance to 29.2 and 29.1 mIoU, respectively.
This highlights the critical role of depth features in our framework and confirms that all four distinct structural complexity descriptors are necessary for the DBC Module.

\begin{table}[ht]
\centering
\caption{Ablation study on the individual components of the total objective.}
\label{tab:ablation-loss}
\begin{tblr}{rowsep=0.5pt, width=\columnwidth, colspec={X[0.75,l,m]X[1,c,m]X[1,c,m]X[1,c,m]X[1,c,m]|X[1,c,m]X[1,c,m]}}
\hline
Exp. & $ \mathcal{L}_\text{con} $ & $ \mathcal{L}_\text{dist} $ & $ \mathcal{L}_\text{align} $ & $ \mathcal{L}_\text{PH} $ & Acc & mIoU \\
\hline
A & \checkmark &  &  &  & 54.5 & 21.7 \\
B & \checkmark & \checkmark &  &  & 59.0 & 22.9 \\
C & \checkmark & \checkmark & \checkmark &  & 64.3 & 26.5 \\
D & \checkmark & \checkmark & \checkmark & \checkmark & 71.1 & 30.2 \\
\hline
\end{tblr}
\end{table}

Second, we quantify the performance gains of each component of our multi-objective loss function, as presented in \Cref{tab:ablation-loss}.
We establish a minimal baseline (A) trained only with the contrastive loss $ \mathcal{L}_\text{con} $, yielding 54.5 Acc and 21.7 mIoU.
By adding the distillation loss $ \mathcal{L}_\text{dist} $ (B), which compares the prototypical cluster outputs of both pathways, consist $ \mathcal{L}_\text{pixel} $, the performance improves by +4.5 Acc and +1.2 mIoU\@.
Then, we add the cross-modal alignment loss $ \mathcal{L}_\text{align} $ (C), which aligns the DBC module's semantic $ f_\text{sem} $ and depth $ f_\text{depth} $ features, causing a substantial performance leap of +5.3 Acc and +3.6 mIoU\@.
Finally, our total objective (D) adds the \lossName~$ \mathcal{L}_\text{PH} $, provides +6.8 Acc and +3.7 mIoU boost to reach our final SOTA performance of 71.1 Acc and 30.2 mIoU\@.

\begin{table}[h]
\centering
\caption{Backbone generalizability analysis against SOTA, including ResNet-152, ConvNeXt, MAE, and DINO.}
\label{tab:ablation-backbone}
\begin{tblr}{rowsep=0.5pt, width=\columnwidth, colspec={X[2.3,l,m]X[1,c,m]X[1,c,m]X[1,c,m]X[1,c,m]}}
\hline
\SetCell[r=2]{} Backbone & \SetCell[c=2]{} Baseline &  & \SetCell[c=2]{}Our Framework \\
& Acc & mIoU & Acc & mIoU \\
\hline
ResNet152 \cite{ResNet} & 9.8 & 3.1 & 15.3 & 8.1 \\
ConvNeXt \cite{ConvNext} & 14.1 & 8.2 & 21.0 & 10.8 \\
SwinViT \cite{SwinV2} & 16.7 & 9.7 & 25.2 & 16.1 \\
MAE \cite{MAE} & 17.4 & 10.1 & 32.4 & 17.8 \\
DINO \cite{DINOv2} & 28.7 & 11.3 & 71.1 & 30.2 \\
\hline
\end{tblr}
\end{table}

Next, we validate the generality of our framework by using different frozen image encoders in our framework, without other modifications. 
The results are shown in \Cref{tab:ablation-backbone}.
Our framework consistently provides a performance lift over the baseline across all tested models.
This holds across different architectures, where our framework improves classic models such as ResNet-152 (+5.0 mIoU) and modern ConvNeXt (+2.6 mIoU) backbones. Crucially, our method demonstrates a +7.7 mIoU gain on the reconstruction-based MAE and a massive +18.9 mIoU gain on the correspondence-based DINO.
The results confirm our framework is a general-purpose structural complexity feature distillation and learning strategy; and structural complexity features hold despite diverse backbones.

\subsection{Computational Cost}
Finally, we analyse the computational cost of our framework to demonstrate that our SOTA performance does not come at the cost of practical efficiency. \Cref{fig:params-flops} provides a comprehensive overview of this trade-off, plotting the mIoU (\%) on COCO-Stuff-27 against the inference-time FLOPs (M) for our method and all other methods analysed in \Cref{subsec:quantitative}. This visualisation clearly shows that both Ours (ViT-S/8) and Ours (ViT-B/8) establish a new performance and efficiency frontier for SSS task.

\section{Conclusion}
\label{sec:conclusion}
In this paper, we identified that existing self-supervised semantic segmentation methods suffer from structural errors because they do not explicitly learn structural complexity features. Even recent depth-fusion approaches, which implicitly encode structure complexity via depth information, fail to resolve structural errors such as merging or fragmentation. Motivated by this observation, we proposed MSSSeg, a framework that explicitly learns multi-scale structural complexity features from both semantic and depth domains through three coupled components: a Differentiable Box-Counting (DBC) module that extracts and aligns multi-scale structural complexity features across modalities, a Learnable Structural Augmentation (StructAug) that forces the network to rely on these features by corrupting pixel-intensity cues, and a Persistent Homology Loss (PHLoss) that directly supervises the structure complexity of the output at critical pixels. Extensive experiments on COCO-Stuff-27, Cityscapes, and Potsdam demonstrate that MSSSeg achieves new state-of-the-art performance without incurring excessive computational overhead. Our results validate that explicitly learning structural complexity features is crucial for robust self-supervised segmentation.
\newpage
%
%
\bibliographystyle{splncs04}
\bibliography{main}

\clearpage
\vspace{1em}
\begin{center}
    \textbf{\Large Supplementary Material}
\end{center}
\setcounter{section}{0}
\renewcommand{\thesection}{\Roman{section}}
\setcounter{figure}{0}
\renewcommand{\thefigure}{\Roman{figure}}
\renewcommand{\thetable}{\Roman{table}}
\renewcommand{\theequation}{\Roman{equation}}
\setcounter{page}{1}

\vspace{0.5em}
\noindent
In the supplementary materials, we provide additional content to support our main manuscript, including:

\begin{enumerate}
    \item Section~\ref{sec:suppl_additional_results} presents additional quantitative and qualitative results on the PASCAL VOC 2012 \cite{PASCALVOC2012}. And qualitative results on Potsdam \cite{Potsdam} datasets.
    
    \item Section~\ref{sec:suppl_pseudo_code} provides the pseudo code for our Differentiable Box Counting (DBC) and Learnable Structural Augmentation (StructAug) modules.
    
    \item Section~\ref{sup_sec:ablation} shows additional ablation studies to validate the design choices of the DBC module, including comparisons between box counting dimension and structural signature representations, as well as comparisons with standard descriptors.
    
    \item Section~\ref{sec:suppl_efficiency} contains detailed quantitative and qualitative analysis of model efficiency.
    
\end{enumerate}

\section{Additional Experiment Results}
\label{sec:suppl_additional_results}

\subsection{Quantitative Analysis}
\begin{table}[ht]
\centering
\caption{Quantitative comparison with SOTA SSS methods on the PASCAL VOC 2012 benchmark. Results are reported for both ViT-S/8 and ViT-B/8 backbones using mIoU. Our framework is highlighted in \colorbox{gray9}{gray} and the best results are in \textbf{bold}.}
\label{tab:quantitative-pascal-voc}
\begin{tblr}{rowsep=0.5pt,
    width=0.7\columnwidth,
    colspec={X[1.5,l,m]X[1,c,m]X[1,c,m]},
    row{6,8} = {gray9}}
\hline
Method & Backbone & mIoU \\
\hline
TransFGU \cite{TransFGU} & ViT-S/8 & 37.2 \\
ACSeg \cite{ACSeg} & ViT-S/8 & 47.1 \\
COMUS \cite{COMUS} & ViT-S/8 & 50.0 \\
CAUSE \cite{CAUSE} & ViT-S/8 & 50.0 \\
\textbf{Ours} & ViT-S/8 & \textbf{55.6} \\
\hline
CAUSE \cite{CAUSE} & ViT-B/8 & 53.3 \\
\textbf{Ours} & ViT-B/8 & \textbf{58.1} \\
\hline
\end{tblr}
\end{table}

We evaluate our framework on the PASCAL VOC 2012 \cite{PASCALVOC2012} dataset to verify its generalization beyond the benchmarks reported in the main paper. Unlike COCO-Stuff and Cityscapes, which focus on dense scene parsing with many overlapping stuff and thing categories, PASCAL VOC contains images dominated by single or few foreground objects against varied backgrounds, testing whether the framework can handle object-centric scenes. As shown in Table~\ref{tab:quantitative-pascal-voc}, our framework outperforms all compared SOTAs across both backbone configurations. With the ViT-S/8 backbone, our framework achieves 55.6\% mIoU, surpassing the strongest baselines COMUS and CAUSE (both at 50.0\%) by 5.6 points. This advantage holds with the larger ViT-B/8 backbone, where our framework reaches 58.1\% mIoU compared to 53.3\% for CAUSE, a gain of 4.8 points. These results confirm that the structural complexity features learned by our framework transfer well to datasets with different scene characteristics and object distributions.

\subsection{Qualitative Analysis}
\label{sup_sec:qualitative}

\begin{figure*}[ht]
    \centering
    \includegraphics[width=0.5\linewidth]{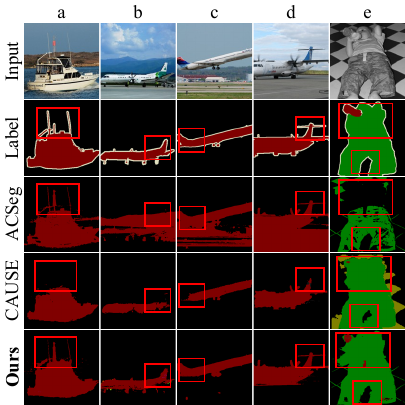}
    \caption{Qualitative segmentation results on PASCAL VOC 2012 \cite{PASCALVOC2012}. Each column (a--e) shows a different image. Rows from top to bottom: input image, ground truth label, ACSeg \cite{ACSeg}, CAUSE \cite{CAUSE}, and our framework. Red boxes mark regions where the compared methods differ most.}
    \label{fig:qualitative-pascal}
\end{figure*}

\begin{figure*}[ht]
    \centering
    \includegraphics[width=0.5\linewidth]{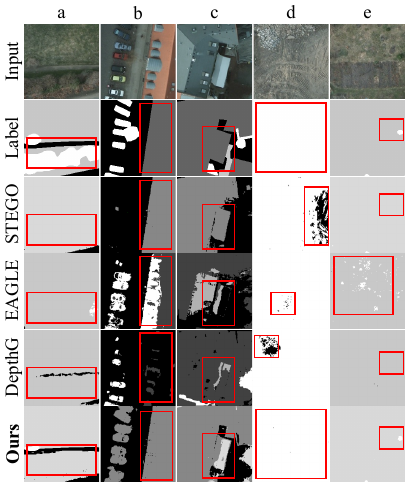}
    \caption{Qualitative segmentation results on Potsdam \cite{Potsdam}. Each column (a--e) shows a different aerial scene. Rows from top to bottom: input image, ground truth label, STEGO \cite{STEGO}, EAGLE \cite{EAGLE}, DepthG \cite{DepthG}, and our framework. Red boxes mark regions where the compared methods differ most.}
    \label{fig:qualitative-potsdam}
\end{figure*}

We present qualitative comparisons on PASCAL VOC 2012 \cite{PASCALVOC2012} in Figure~\ref{fig:qualitative-pascal} and on Potsdam \cite{Potsdam} in Figure~\ref{fig:qualitative-potsdam}. Red boxes in both figures highlight regions where SOTAs show notable differences.

Figure~\ref{fig:qualitative-pascal} compares our framework with ACSeg \cite{ACSeg} and CAUSE \cite{CAUSE} on five PASCAL VOC images. In Column a (boat), ACSeg produces a noisy mask with background regions leaking into the boat, and CAUSE leaves parts of the mast missing. Our framework segments the full boat structure, matching the ground truth more closely. In Columns b, c, and d (planes). ACSeg fails to segment the plane from shadow, while CAUSE captures the overall shape but shows artifacts near the tail. Our result recovers the tail region cleanly. In Column e (baby on a checkered floor with a bottle), ACSeg confuses the checkered background texture with the foreground, introducing scattered green labels. CAUSE shows yellow artifacts leaking into the lower region. Our framework separates the person from the textured background with fewer errors.

Figure~\ref{fig:qualitative-potsdam} compares our framework with STEGO \cite{STEGO}, EAGLE \cite{EAGLE}, and DepthG \cite{DepthG} on five aerial scenes. In Column a (road through vegetation), STEGO assigns the road to the wrong class, making it indistinguishable from the surrounding areas. EAGLE also misclassifies the road region. DepthG produces a thin, fragmented road segment. Our framework correctly identifies the road and separates it from the vegetation, consistent with the ground truth. In Column b (buildings with parked cars), STEGO merges the roof with the background. EAGLE oversegments the scene, producing scattered white blobs across the buildings. DepthG fragments the roof boundary. Our framework produces clean building outlines that match the label. In Column c (rooftops), STEGO misses building details in this region, and EAGLE introduces noisy fragments. DepthG produces artifacts where the building meets the background. Our result preserves the building structure accurately. In Columns d and e (ground and field scenes), STEGO and EAGLE introduce noisy patches and false detections. Our framework produces uniform segmentation maps without these artifacts, closely matching the ground truth labels.

By jointly leveraging multi-scale structural complexity features, learnable structural augmentations, and topological constraints, our framework produces sharper boundaries and fewer fragmented regions than the compared methods.

\section{Pseudo Code for Proposed Modules}
\label{sec:suppl_pseudo_code}

We provide the pseudo-code for the two proposed modules described in the main paper. Algorithm~\ref{alg:dbc} details the DBC module (Section 3.2 of the main paper), which extracts structural complexity signatures from both the depth map and the pseudo label mask, projects them through separate MLPs, and computes the alignment loss. Algorithm~\ref{alg:structaug} details the StructAug module (Section 3.3 of the main paper), which converts the input image to HSL, applies learnable geodesic morphological operators to the Lightness channel, selects among them via Gumbel-Softmax sampling, and blends the result back to produce the augmented image.

\begin{algorithm}[t]
\caption{DBC Module}
\label{alg:dbc}
\begin{algorithmic}[1]
\Require Pseudo-label map $M_\text{sem} \in \mathbb{R}^{B \times H \times W}$; normalized depth map $M_\text{depth} \in \mathbb{R}^{B \times H \times W}$; box sizes $\mathbf{S} = \{s_1, \ldots, s_k\}$; depth MLP $\phi_\text{depth}$; semantic MLP $\phi_\text{sem}$
\Ensure Structural feature $f_\text{struct} \in \mathbb{R}^{B \times 2D}$; alignment loss $\mathcal{L}_\text{align}$

\Statex \hspace{\algorithmicindent}\textit{// Core differentiable box counting (Eq.~4,~5 of the main paper)}
\Function{DBC}{$M, \mathbf{S}$}
    \For{each $s \in \mathbf{S}$}
        \State $\DBC(M, s) \gets \sum_{i,j} \maxpool_{s \times s}(M)[i,j]$ \Comment{Continuous occupancy}
    \EndFor
    \State \Return $\bigl(\DBC(M, s)\bigr)_{s \in \mathbf{S}}$ \Comment{$k$-dim complexity signature}
\EndFunction

\Statex
\Statex \hspace{\algorithmicindent}\textit{// Depth branch: surface roughness + depth edges}
\State $M_\text{depth\_var} \gets \mathbb{E}_{3\times3}[M_\text{depth}^2] - \mathbb{E}_{3\times3}[M_\text{depth}]^2$ \Comment{Local variance map}
\State $N_\text{depth\_rough} \gets \Call{DBC}{M_\text{depth\_var},\; \mathbf{S}}$ \Comment{Surface roughness descriptor}
\State $G_\text{depth} \gets \text{Sobel}(M_\text{depth})$ \Comment{Depth gradient magnitude}
\State $N_\text{depth\_bound} \gets \Call{DBC}{G_\text{depth},\; \mathbf{S}}$ \Comment{Depth edge descriptor}

\Statex
\Statex \hspace{\algorithmicindent}\textit{// Semantic branch: area counting + boundary counting}
\State $M_\text{sem} \gets \text{OneHot}(M_\text{sem}) \in \{0,1\}^{B \times C \times H \times W}$ \Comment{Per-class binary masks}
\State $N_\text{sem\_area} \gets \Call{DBC}{M_\text{sem},\; \mathbf{S}}$ \Comment{Area counting descriptor}
\State $G_\text{seg} \gets \text{Dilate}(M_\text{sem}) - \text{Erode}(M_\text{sem})$ \Comment{Morphological gradient}
\State $N_\text{sem\_bound} \gets \Call{DBC}{G_\text{seg},\; \mathbf{S}}$ \Comment{Boundary counting descriptor}

\Statex
\Statex \hspace{\algorithmicindent}\textit{// Domain-specific projection and alignment}
\State $f_\text{depth} \gets \phi_\text{depth}\bigl([N_\text{depth\_rough};\; N_\text{depth\_bound}]\bigr) \in \mathbb{R}^{B \times D}$ \Comment{Depth MLP}
\State $f_\text{sem} \gets \phi_\text{sem}\bigl([N_\text{sem\_area};\; N_\text{sem\_bound}]\bigr) \in \mathbb{R}^{B \times D}$ \Comment{Semantic MLP}
\State $\mathcal{L}_\text{align} \gets 1 - \cos(f_\text{sem},\; f_\text{depth})$ \Comment{Cross-modal alignment loss}
\State $f_\text{struct} \gets [f_\text{sem};\; f_\text{depth}] \in \mathbb{R}^{B \times 2D}$ \Comment{Concatenated structural feature}
\State \Return $f_\text{struct}$, $\mathcal{L}_\text{align}$
\end{algorithmic}
\end{algorithm}

\begin{algorithm}[t]
\caption{\augName}
\label{alg:structaug}
\begin{algorithmic}[1]
\Require Image $I$; learnable parameters $\theta = \{\sigma_\varepsilon, \sigma_\delta, \alpha_\varepsilon, \alpha_\delta, \{\ell_i\}_{i=1}^{4}, \ell_\beta\}$; neighborhood size $k$; Gumbel temperature $\tau$
\Ensure Augmented image $I_{\text{aug}}$
\State $(H, S, L) \gets \texttt{RGB2HSL}(I)$ \Comment{Convert to HSL, isolate Lightness}
\Statex \hspace{\algorithmicindent}\textit{// Learnable geodesic erosion and dilation}
\For{each pixel $x$}
    \For{each $y \in \mathcal{N}_k(x)$}
        \State $w_\theta(x, y) \gets \exp\!\bigl(-{|L(x) - L(y)|^2}\big/{2\,\texttt{softplus}(\sigma_\theta)^2}\bigr)$ \Comment{Affinity kernel}
    \EndFor
    \State $\varepsilon_\theta(L)(x) \gets \min_{y \in \mathcal{N}_k(x)} \bigl[ L(y) + (1 - w_\theta(x,y)) \cdot \texttt{softplus}(\alpha_\varepsilon) \bigr]$ \Comment{Geodesic erosion}
    \State $\delta_\theta(L)(x) \gets \max_{y \in \mathcal{N}_k(x)} \bigl[ L(y) - (1 - w_\theta(x,y)) \cdot \texttt{softplus}(\alpha_\delta) \bigr]$ \Comment{Geodesic dilation}
\EndFor
\Statex \hspace{\algorithmicindent}\textit{// Compose morphological operators}
\State $\mathcal{O}_\theta(L) \gets \delta_\theta\bigl(\varepsilon_\theta(L)\bigr)$ \Comment{Opening: suppresses bright bridges}
\State $\mathcal{C}_\theta(L) \gets \varepsilon_\theta\bigl(\delta_\theta(L)\bigr)$ \Comment{Closing: fills dark gaps}
\State $\mathcal{B}_\theta(L) \gets \mathcal{C}_\theta(L) - L$ \Comment{Black Top-Hat: shadow residuals}
\State $\mathcal{W}_\theta(L) \gets L - \mathcal{O}_\theta(L)$ \Comment{White Top-Hat: glare residuals}
\State $\mathcal{T} \gets \{\mathcal{O}_\theta(L),\; \mathcal{C}_\theta(L),\; \mathcal{B}_\theta(L),\; \mathcal{W}_\theta(L)\}$ \Comment{Operator toolbox}
\Statex \hspace{\algorithmicindent}\textit{// Gumbel-Softmax operator selection}
\For{$i = 1, \ldots, |\mathcal{T}|$}
    \State $g_i \sim \text{Gumbel}(0, 1)$
    \State $\tilde{p}_i \gets \exp\bigl((\ell_i + g_i) / \tau\bigr) \big/ \sum_{j} \exp\bigl((\ell_j + g_j) / \tau\bigr)$
\EndFor
\State $\hat{T} \gets \sum_{i} \tilde{p}_i \cdot T_i$ \Comment{Differentiable soft selection}
\Statex \hspace{\algorithmicindent}\textit{// Learnable blending}
\State $\beta \gets \text{sigmoid}(\ell_\beta)$ \Comment{Blending coefficient}
\State $L_{\text{aug}} \gets (1 - \beta)\, L + \beta\, \hat{T}$ \Comment{Blend original and corrupted}
\State $I_{\text{aug}} \gets \texttt{HSL2RGB}(H, S, L_{\text{aug}})$ \Comment{Reconstruct augmented image}
\State \Return $I_{\text{aug}}$
\end{algorithmic}
\end{algorithm}

\section{Additional Ablation Study}
\label{sup_sec:ablation}
\subsection{Box Sizes of DBC Module}

We analyze the sensitivity of the DBC module to its main hyperparameter: the set of box sizes $\mathbf{S} = \{s_1, s_2, \ldots, s_k\}$. All experiments in this section use the COCO-Stuff-27 benchmark with the DINO pretrained ViT-S/8 backbone and report Acc and mIoU.

Table~\ref{tab:ablation-hyperparameters} shows results for seven configurations of box sizes. Two findings stand out.

\begin{table}[ht]
\centering
\caption{Ablation on the set of box sizes $\mathbf{S}$ for the DBC module. Our selected configuration is highlighted in \colorbox{gray9}{gray}. Best results are in \textbf{bold}.}
\label{tab:ablation-hyperparameters}
\begin{tblr}{rowsep=0.5pt,
    width=\columnwidth,
    colspec={X[1,c,m]X[2,l,m]X[1,c,m]},
    row{7} = {gray9}}
\hline
Exp. & Box sizes & Acc & mIoU\\
\hline
a. & $ \{ 4,8,16 \} $ & 66.4 & 24.3 \\
b. & $ \{ 8,16,32 \} $ & 65.7 & 25.7 \\
c. & $ \{ 16,32,64 \} $ & 66.8 & 25.4 \\
d. & $ \{ 32,64,128 \} $ & 66.5 & 24.7 \\
e. & $ \{ 4,8,16,32 \} $ & 67.1 & 26.8 \\
f. & $ \{ 2,4,8,16,32,64 \} $ & \textbf{71.1} & \textbf{30.2} \\
g. & $ \{ 2,4,8,16,32,64,128 \} $ & 69.7 & 29.1 \\
\hline
\end{tblr}
\end{table}

First, a wider range of box sizes leads to better performance. Configurations that span only a narrow range (configs a through d, each with three sizes) all score below 26.0 mIoU. Adding more sizes helps: config e (four sizes) reaches 26.8, and config f (six sizes, from $s{=}2$ to $s{=}64$) reaches \textbf{30.2}. This is consistent with the principle behind box counting: the structural complexity signature is defined by how occupancy changes across scales, so covering both small and large scales produces a more informative descriptor.

Second, performance drops when the largest box size becomes too large relative to the image resolution. Config d ($\{32, 64, 128\}$) scores only 24.7 mIoU, and config g, which adds $s{=}128$ to the best config f, drops from 30.2 to 29.1. At a training resolution of $224 \times 224$, a box of size 128 covers more than a quarter of the image. The resulting box counts are too coarse to be discriminative and instead introduce noise into the projection head. Based on these results, we select config f as our default setting.

\subsection{Box Counting Dimension vs. Structural Signature}
\label{sec:ablation_dimension}

The DBC module is motivated by the box counting dimension from fractal geometry~\cite{mandelbrot1982fractal}. The classical definition computes a single scalar $D$ as the slope of $\log N(s)$ versus $\log(1/s)$:
\begin{equation}
    D = \lim_{s \to 0} \frac{\log N(s)}{\log (1/s)}
\end{equation}
where $N(s)$ is the number of boxes of size $s$ needed to cover the structure. In practice, $D$ is estimated by fitting a line to the log-log plot via least squares regression. However, this scalar assumes that occupancy varies uniformly across scales, which holds for ideal fractals but not for natural scenes. Object boundaries in real images have different characteristics at different scales: a building edge may appear smooth at coarse scales but irregular at fine scales, while a tree canopy may show the opposite pattern. Our DBC module therefore keeps the full vector of occupancy counts across $K$ scales, forming a $K$ dimensional signature rather than collapsing it into a single scalar. This preserves per-scale structural differences and lets the projection MLPs weight each scale independently, which is important for the alignment between the semantic and depth branches since the two domains may share structure at some scales but not others.

To verify this benefit, we modify the DBC module to compute the scalar slope $D$ via differentiable linear regression during the forward pass. This scalar replaces the $K$ dimensional signature for both the alignment loss and the downstream projection.

\begin{table}[h]
\centering
\caption{Scalar box counting dimension vs. our $K$ dimensional signature on COCO-Stuff-27 (ViT-S/8).}
\label{tab:strict_dimension}
\begin{tabular}{lcc}
\toprule
Feature Representation & Acc (\%) & mIoU (\%) \\
\midrule
Scalar Box Counting Dimension (Slope) & 65.1 & 22.8 \\
\textbf{DBC Signature ($K$ dim, Ours)} & \textbf{71.1} & \textbf{30.2} \\
\bottomrule
\end{tabular}
\end{table}

As shown in Table~\ref{tab:strict_dimension}, collapsing the signature into a single scalar reduces mIoU from 30.2 to 22.8, a drop of 7.4 points. This confirms that the per-scale structural information captured by the full signature is essential and cannot be summarized by a single slope value.

\subsection{Multi-Scale Design vs. Box Counting Mechanism}
\label{sec:ablation_multiscale}

The previous subsection shows that our $K$ dimensional signature outperforms a scalar dimension. To further verify that both the box counting mechanism and the multi-scale design contribute to the final performance, we design three baselines to disentangle the two factors. First, we reduce the DBC module to a single box size ($ s = 64 $) to isolate the effect of multi-scale coverage. Second, we replace the DBC max pooling operation (Equation 4 of the main paper) with average pooling at the same $K$ scales, which preserves the multi-scale design but removes the occupancy counting that is specific to box counting. Third, we build a $K$ level Gaussian pyramid from the input feature maps and compute variance and gradient magnitude at each level, which captures multi-scale frequency content without any box counting operation.

\begin{table}[h]
\centering
\caption{Comparison of the DBC mechanism with standard descriptors on COCO-Stuff-27 (ViT-S/8). The first row uses DBC at a single scale to isolate the multi-scale factor. The next two rows use multi-scale features without the box counting operation.}
\label{tab:multiscale_baselines}
\begin{tabular}{lcc}
\hline
Descriptor & Acc (\%) & mIoU (\%) \\
\hline
Single Scale DBC & 60.1 & 19.6 \\
\hline
Multi-Scale Average Pooling & 59.7 & 19.4 \\
Multi-Scale Gaussian Pyramid & 64.1 & 22.8 \\
\textbf{Multi-Scale DBC Signature (Ours)} & \textbf{71.1} & \textbf{30.2} \\
\hline
\end{tabular}
\end{table}

As shown in Table~\ref{tab:multiscale_baselines}, single scale DBC scores 19.6 mIoU, which is between two multi-scale alternatives that do not use box counting (19.4 and 22.8), showing that the box counting operation itself provides a strong structural signal even at one scale. However, it falls well below the full multi-scale DBC (30.2), confirming that covering multiple scales is equally important. Among the multi-scale baselines, average pooling smooths over spatial discontinuities and cannot distinguish a single connected region from several scattered fragments with the same total activation. Gaussian pyramids capture frequency content at each level but do not measure spatial occupancy directly. The box counting mechanism, by contrast, counts how many spatial cells are occupied at each scale, which directly encodes connectivity, fragmentation, and boundary complexity. These results confirm that the performance of our framework comes from the combination of the box counting operation and the multi-scale design, not from either factor alone.

\section{Details of Model Efficiency}
\label{sec:suppl_efficiency}

\begin{table*}[!t]
\centering
\caption{Efficiency comparison on COCO-Stuff-27. Param is the number of parameters in millions. FLOPs is the computation cost at inference in millions. Our framework is highlighted in \colorbox{gray!30}{gray} and the best mIoU is in \textbf{bold}.}
\label{tab:efficiency}
\begin{tblr}{rowsep=0.5pt,
    width=\linewidth,
    colspec={X[2,l,m]X[1.5,c,m]X[1,c,m]X[1,c,m]X[1,c,m]},
    row{10,15} = {gray9}}
\hline
Method & Backbone & Param & FLOPs & mIOU \\
\hline
DINO & ViT-S/8 & 22.1 & 68.3 & 11.3 \\
STEGO & ViT-S/8 & 21.9 & 268.2 & 24.5 \\
TransFGU & ViT-S/8 & 24.4 & 19.0 & 17.5 \\
ACSeg & ViT-S/8 & 23.8 & 17.9 & 16.4 \\
HP & ViT-S/8 & 22.3 & 17.1 & 24.3 \\
PriMaPs & ViT-S/8 & 21.8 & 16.9 & 16.4 \\
DepthG & ViT-S/8 & 21.9 & 280.2 & 26.7 \\
EAGLE & ViT-S/8 & 25.1 & 2077.6 & 27.2 \\
\textbf{Ours} & ViT-S/8 & 24.2 & 279.3 & \textbf{30.2} \\
\hline
DINO & ViT-B/8 & 86.6 & 133.4 & 9.6 \\
STEGO & ViT-B/8 & 86.6 & 761.9 & 28.2 \\
PriMaPs & ViT-B/8 & 85.9 & 67.0 & 21.9 \\
DepthG & ViT-B/8 & 86.6 & 774.0 & 29.0 \\
\textbf{Ours} & ViT-B/8 & 91.8 & 643.9 & \textbf{34.9} \\
\hline
\end{tblr}
\end{table*}

Along with the efficiency discussion in our main paper, we present detailed metrics for all compared methods on COCO-Stuff-27 in Table~\ref{tab:efficiency}.

With the ViT-S/8 backbone, our framework uses 24.2M parameters and 279.3M FLOPs, reaching 30.2 mIoU. This is comparable in cost to DepthG (21.9M parameters, 280.2M FLOPs) but 3.5 points higher in mIoU. EAGLE reaches 27.2 mIoU but requires 2077.6M FLOPs, roughly 7.4 times more computation than ours. Lighter methods such as HP (17.1M FLOPs), ACSeg (17.9M FLOPs), and PriMaPs (16.9M FLOPs) use far fewer FLOPs but score 24.3, 16.4, and 16.4 mIoU, respectively, well below our result.

With the ViT-B/8 backbone, our framework reaches 34.9 mIoU with 91.8M parameters and 643.9M FLOPs. DepthG uses a similar number of FLOPs (774.0M) but scores 29.0, a gap of 5.9 points. STEGO uses 761.9M FLOPs and scores 28.2. Our framework achieves the best performance under both backbones while maintaining comparable computational cost to methods in the same accuracy range.
\end{document}